\documentclass[letterpaper, 10 pt, conference]{ieeeconf}

\IEEEoverridecommandlockouts
\usepackage{graphicx}
\usepackage{array, makecell}
\usepackage{multirow}

\title{\LARGE Traffic-Aware Navigation in Road Networks}

\author{Sarah Nassar$^{1}$
\thanks{$^{1}$Sarah Nassar is with the Department of Electrical and Computer Engineering at Queen's University in Kingston, Ontario, Canada}%
}

\begin{document}

\maketitle
\thispagestyle{empty}
\pagestyle{empty}


\begin{abstract}
This project compares three graph search approaches for the task of traffic-aware navigation in Kingston's road network. These approaches include a single-run multi-query preprocessing algorithm (Floyd-Warshall-Ingerman), continuous single-query real-time search (Dijkstra's and A*), and an algorithm combining both approaches to balance between their trade-offs by first finding the top K shortest paths then iterating over them in real time (Yen's). Dijkstra's and A* resulted in the most traffic-aware optimal solutions with minimal preprocessing required. Floyd-Warshall-Ingerman was the fastest in real time but provided distance based paths with no traffic awareness. Yen's algorithm required significant preprocessing but balanced between the other two approaches in terms of runtime speed and optimality. Each approach presents advantages and disadvantages that need to be weighed depending on the circumstances of specific deployment contexts to select the best custom solution. \textit{*This project was completed as part of ELEC 844 (Search and Planning Algorithms for Robotics) in the Fall 2025 term.}
\end{abstract}



\section{INTRODUCTION}

Road navigation, whether through Global Positioning System (GPS) devices or mobile applications, has become an essential and ubiquitous utility in modern life. The core technology facilitating navigation tools is graph search, since road networks can intuitively be represented as graphs with intersections as vertices and roads as edges. The primary goal is to find an optimal route to drive between two locations on a map using public roads. Optimality is typically defined in terms of time, so the objective is to arrive at the destination as fast as possible. To accomplish this, factors such as distance, speed limits, and traffic conditions play a key role in defining the cost of edges.

An important practical consideration is how fast a low-cost route can be found on-demand. Therefore, the purpose of this project is to compare several path-finding approaches across multiple metrics that encompass their performance from various perspectives, including preprocessing time, runtime, and path cost.

The environment of interest is Kingston's road network, but the algorithms tested here can be used for any city as long as its road network can be easily represented as a graph. (Navigating between cities or within larger cities may require different algorithms or planning paradigms that combine global and local planning.) The task does not revolve around a specific robot, but it can be used as a global planner for an autonomous vehicle or it could be used to provide instructions to a human driver. Since the approach is to implement global planners, obstacles are not considered. This is an interesting problem since a city's road network is large, so implementing feasible and efficient algorithms is challenging.


\section{RELATED WORK}

There exist several popular navigation apps, most notably Google Maps which integrates traffic data. Real-time traffic information can be collected from various sources, such as crowdsourced data from active users or government-sourced data from sensors and cameras embedded in roads, traffic lights, or highways \cite{cohn2009real}. There are several studies that explore traffic-aware routing. Two such examples are \cite{xu2012traffic} and \cite{shang2013finding}, but a limitation is that they focus on single-query search (e.g., Dijkstra's algorithm or A* search) without considering offloading some of the computational effort to a one-time preprocessing procedure. Therefore, to our knowledge, this is the first study to propose a K shortest paths approach and include it in a comparison with the more traditional single-query search approach.


\section{METHODS}

\subsection{Data Preparation}

\begin{figure}
    \centering
    \includegraphics[width=\linewidth]{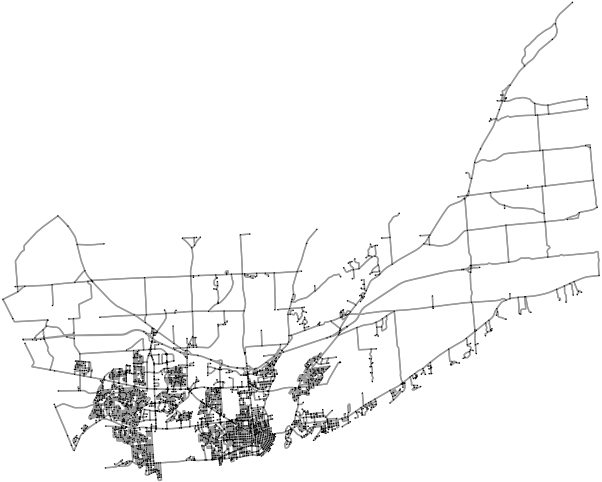}
    \caption{Graph representation of Kingston's road network, downloaded and plotted with the OSMnx Python package.}
    \label{fig:osmnxmap}
\end{figure}

\begin{figure}
    \centering
    \includegraphics[width=\linewidth]{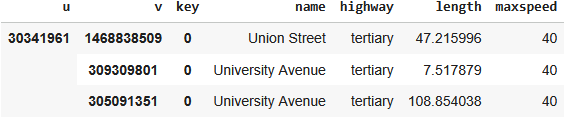}
    \caption{Subset of rows and columns from edges DataFrame.}
    \label{fig:df}
\end{figure}

The OSMnx Python package was used to download open-source road network data \cite{boeing2025modeling}. For Kingston, the graph included 3,305 vertices (i.e., intersections) and 8,467 directed edges (i.e., road segments connecting intersections). The majority of these road segments were residential (n=5,447) and most of them had missing speed limit information (n=6,689). Fig. \ref{fig:osmnxmap} depicts Kingston's road network, while Fig. \ref{fig:df} shows a subset of rows and columns from the edges DataFrame.

There were 200 pairs of parallel edges. A parallel edge signified two possible road segments that can be taken to move between two adjacent intersections. The approach to handle parallel edges was to choose the edge minimizing the relevant edge cost (i.e., distance for preprocessing or traffic-weighted distance for real-time search).

The 'maxspeed' column in the edges DataFrame had two major issues: 1) most speed limits were missing (n=6,689) and 2) some edges (i.e., road segments) had multiple speed limits (n=48). To address the missing speed limits, values were imputed based on the road type. The speed limit was assumed to be 50 km/h if the road type was 'residential', 'primary', 'unclassified', 'motorway\_link', 'secondary\_link', 'primary\_link', or 'tertiary\_link', 80 km/h if the road type was 'secondary' or 'tertiary', and 100 km/h if the road type was 'motorway'. If the edge contained multiple road types or speed limits, the average was used.

\subsection{Algorithms}

Three graph search approaches were included in the comparative experiments:
\begin{enumerate}
    \item \textbf{Multi-Query Lookup Approach:} The first approach uses the Floyd-Warshall-Ingerman algorithm \cite{warshall1962theorem, floyd1962algorithm, ingerman1962algorithm} using distances only as edge costs. This is a multi-query algorithm that finds the optimal path between every pair of vertices on the graph. The preprocessing step takes a notable amount of time, but having a lookup table will result in fast real-time results. The main downside is that this method would only optimize path distance, without taking into consideration real-time traffic conditions that may affect arrival time since the lowest-distance path may have higher traffic and take more time than a longer path with low traffic.
    \item \textbf{Single-Query Search Approach:} The second approach is to run a single-query Dijkstra's algorithm \cite{dijkstra1959note} or A* search \cite{hart1968formal} each time a path is requested, with up-to-date traffic information integrated with the edge costs. This will result in traffic-optimized solutions, but the main downside is that a new search must be conducted upon each user request, which may result in more real-time processing and longer wait times. Multiple search heuristics were tested, including the zero heuristic (i.e., Dijkstra's algorithm), Euclidean distance, great-circle distance, and inflated heuristics (which result in faster searches but sub-optimal solutions).
    \item \textbf{K Shortest Paths Approach:} The third algorithm is to use Yen's algorithm \cite{yen1971finding} to find the K shortest paths between each pair of vertices, with distances as edge costs. This algorithm finds the top-K optimal loopless paths between two vertices, and can be used in a preprocessing step. Then, the real-time task would be to iterate over these K candidate paths to calculate the traffic-weighted path cost and choose the path minimizing this cost. The solutions should be more optimal than the first multi-query lookup approach and faster in real time than the second single-query search approach. However, the preprocessing step takes a significant amount of time. In this project, K=5.
\end{enumerate}

All three algorithms were implemented for this project using Python. OSMnx provided access to existing implementations for Dijkstra's algorithm (without traffic integration) and a K shortest paths algorithm. These were tested to verify the correctness of our implementations. The OSMnx implementations were faster, but provided the same final average costs as our implementations.

\subsection{Metrics}

Three metrics were used to compare algorithms. These are:
\begin{itemize}
    \item \textbf{Preprocessing Overhead:} Total time required to run algorithm initialization.
    \item \textbf{Real-Time Computation Effort:} Average runtime.
    \item \textbf{Optimality:} Average solution cost (i.e., traffic-weighted distance).
\end{itemize}

For Approach \#2 (Single-Query Search), the number of vertices expanded will also be reported.


\subsection{Experimental Setup}

\begin{figure*}
    \centering
    \includegraphics[width=\textwidth]{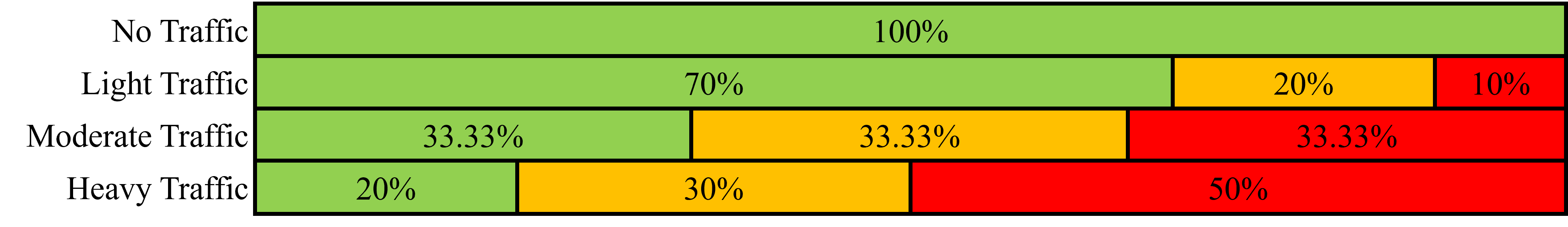}
    \caption{Probability distribution of low (green), medium (orange), and high (red) traffic weights for each set of trials.}
    \label{fig:traffic}
\end{figure*}

\begin{figure}
    \centering
    \includegraphics[width=\linewidth]{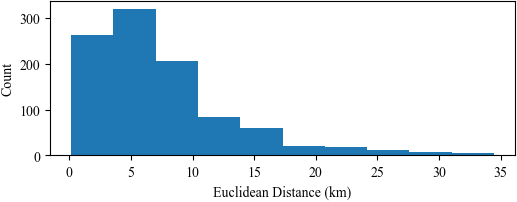}
    \caption{Histogram of Euclidean distance between each of the 1,000 start-goal intersection pairs.}
    \label{fig:hist1}
\end{figure}

1,000 trials were run as part of this experiment. These were generated by randomly sampling 1,000 start-goal intersection pairs (without replacement so all 2,000 vertices are unique), and assigning a set of traffic weights for the edges in each trial. Traffic is classified as low, medium, and high, with weights 1, 3, and 5, respectively. 250 trials had no traffic (i.e., all edges were assigned weight=1), 250 trials had light traffic (i.e., the probability distribution for sampling traffic weights was 70\% for weight=1, 20\% for weight=3, and 10\% for weight=5), 250 trials had moderate traffic (i.e., the probability distribution for sampling traffic weights was uniform across all three weights), and 250 trials had heavy traffic (i.e., the probability distribution for sampling traffic weights was 20\% for weight=1, 30\% for weight=3, and 50\% for weight=5). Fig. \ref{fig:traffic} presents these traffic conditions in visual form. Fig. \ref{fig:hist1} plots a histogram of the Euclidean distance between each of the 1,000 start-goal intersection pairs. For reproducibility, the seed for NumPy's random number generator was set to 0.

\section{RESULTS}

Out of the 1,000 trials, five did not have a solution (i.e., there was no possible path between the start and goal intersections).

\subsection{Traffic-Weighted Distance as Cost}

Table \ref{tab:results1} shows the experiment metrics for all algorithms, with traffic-weighted distances as the edge costs. In terms of preprocessing time, Dijkstra's algorithm is the only one that did not require any preprocessing. A* search and inflated A* required minimal preprocessing in under 20 minutes to generate heuristic lookup tables (i.e., the Euclidean or great-circle distance between each pair of vertices). The Floyd-Warshall-Ingerman algorithm completed preprocessing in over 4.5 hours. Yen's algorithm took the longest time for preprocessing, which required almost 66 hours. Due to the significant amount of time to run the preprocessing for Yen's algorithm, the K shortest paths were only computed for the 1,000 start-goal pairs used as experimental trials. For actual deployment, the preprocessing would have to be executed for all pairs of vertices. At the rate of our implementation, this would not be feasible. Even with OSMnx's faster implementation, the complete preprocessing is estimated to take about 45 days for K=5.

In terms of average runtime, the fastest algorithm is the Floyd-Warshall-Ingerman algorithm, followed by Yen's algorithm, then the inflated A* algorithms, and finally Dijkstra's algorithm and A* search. As expected, offloading all or most of the computation to the preprocessing stage for the Floyd-Warshall-Ingerman algorithm and Yen's algorithm results in minimal runtime computation.

In terms of average cost, the optimal paths were found by Dijkstra's algorithm and A* (regardless of heuristic). The next best paths were found by Yen's algorithm, followed by the Floyd-Warshall-Ingerman algorithm. The worst paths were found by the inflated A* searches.

In terms of the average number of vertices expanded, Dijkstra's algorithm had to process the most vertices, followed by A* with the great-circle distance heuristic, then A* with Euclidean distance, then inflated A* with $\alpha$=10, and finally inflated A* with $\alpha$=100. Although A*, as a heuristic-based informed search, expanded less nodes than Dijkstra's algorithm, this did not result in a significant different in runtime. A potential explanation could be that performing heuristic lookups added an extra real-time computational overhead.

\begin{table*}
\caption{Performance metrics for all algorithms across the three approaches with traffic-weighted distance as edge cost. **Only partial preprocessing was conducted for Yen's algorithm.}
\label{tab:results1}
\centering
\begin{tabular}{|c|c|c|c|c|c|}
\hline
Approach & Algorithm & \makecell[c]{Preprocessing \\ Time (min)} & \makecell[c]{Average \\ Runtime (s)} & \makecell[c]{Average \\ Cost (km)} & \makecell[c]{Average \# of \\ Vertices Expanded} \\ \hline
Multi-Query Lookup & Floyd-Warshall-Ingerman & 273.01 & \textbf{0.000040} & 21.78 & - \\ \hline
\multirow{7}{*}{Single-Query Search}
 & Dijkstra's & \textbf{0.00} & 2.38 & \textbf{19.31} & 1632 \\ \cline{2-6}
 & A* (Euclidean) & 15.63 & 2.12 & \textbf{19.31} & 997 \\ \cline{2-6}
 & A* (Great-Circle) & 18.57 & 2.38 & \textbf{19.31} & 998 \\ \cline{2-6}
 & Inflated A* (Euclidean, $\alpha$=10) & 15.63 & 0.20 & 23.52 & 64 \\ \cline{2-6}
 & Inflated A* (Euclidean, $\alpha$=100) & 15.63 & 0.18 & 25.46 & \textbf{63} \\ \cline{2-6}
 & Inflated A* (Great-Circle, $\alpha$=10) & 18.57 & 0.20 & 23.49 & 64 \\ \cline{2-6}
 & Inflated A* (Great-Circle, $\alpha$=100) & 18.57 & 0.19 & 25.46 & \textbf{63} \\ \hline
K Shortest Paths & Yen's & 3946.91** & 0.043 & 21.36 & - \\ \hline
\end{tabular}
\end{table*}

Table \ref{tab:results2} shows a comparison of the average path length, cost, and estimated arrival time for all algorithms, with traffic-weighted distances as the edge costs. Path length is computed as the sum of all edge distances part of the solution, cost is computed as the sum of traffic-weighted distances (i.e., traffic weight multiplied by distance), and estimated arrival time is computed as the sum of traffic and speed limit-weighted distances (i.e., traffic multiplied by distance divided by speed limit). Although vehicles cannot move at the speed limit in congested traffic, the effect of applying a traffic weight can be intuitively understood as increasing the distance or reducing the speed. It can be noted that while the average path length for the Floyd-Warshall-Ingerman algorithm is shorter than it is for Dijkstra's algorithm, A* search, and Yen's algorithm, it has a higher average cost given traffic conditions and a longer estimated arrival time.

\begin{table*}
\caption{Comparison of path length, traffic-weighted costs, and estimated arrival time across algorithms with traffic-weighted distance as edge cost.}
\label{tab:results2}
\centering
\begin{tabular}{|c|c|c|c|c|}
\hline
Approach & Algorithm & \makecell[c]{Average \\ Length (km)} & \makecell[c]{Average \\ Cost (km)} & \makecell[c]{Average Estimated \\ Arrival Time (min)} \\ \hline
Multi-Query Lookup & Floyd-Warshall-Ingerman & \textbf{9.12} & 21.78 & 21.89 \\ \hline
\multirow{7}{*}{Single-Query Search}
 & Dijkstra's & 9.92 & \textbf{19.31} & \textbf{19.58} \\ \cline{2-5}
 & A* (Euclidean) & 9.92 & \textbf{19.31} & \textbf{19.58} \\ \cline{2-5}
 & A* (Great-Circle) & 9.92 & \textbf{19.31} & \textbf{19.58} \\ \cline{2-5}
 & Inflated A* (Euclidean, $\alpha$=10) & 10.31 & 23.52 & 23.83 \\ \cline{2-5}
 & Inflated A* (Euclidean, $\alpha$=100) & 10.84 & 25.46 & 25.74 \\ \cline{2-5}
 & Inflated A* (Great-Circle, $\alpha$=10) & 10.31 & 23.49 & 23.80 \\ \cline{2-5}
 & Inflated A* (Great-Circle, $\alpha$=100) & 10.84 & 25.46 & 25.74 \\ \hline
K Shortest Paths & Yen's & 9.15 & 21.36 & 21.44 \\ \hline
\end{tabular}
\end{table*}

\begin{figure*}
    \centering
    \includegraphics[width=\textwidth]{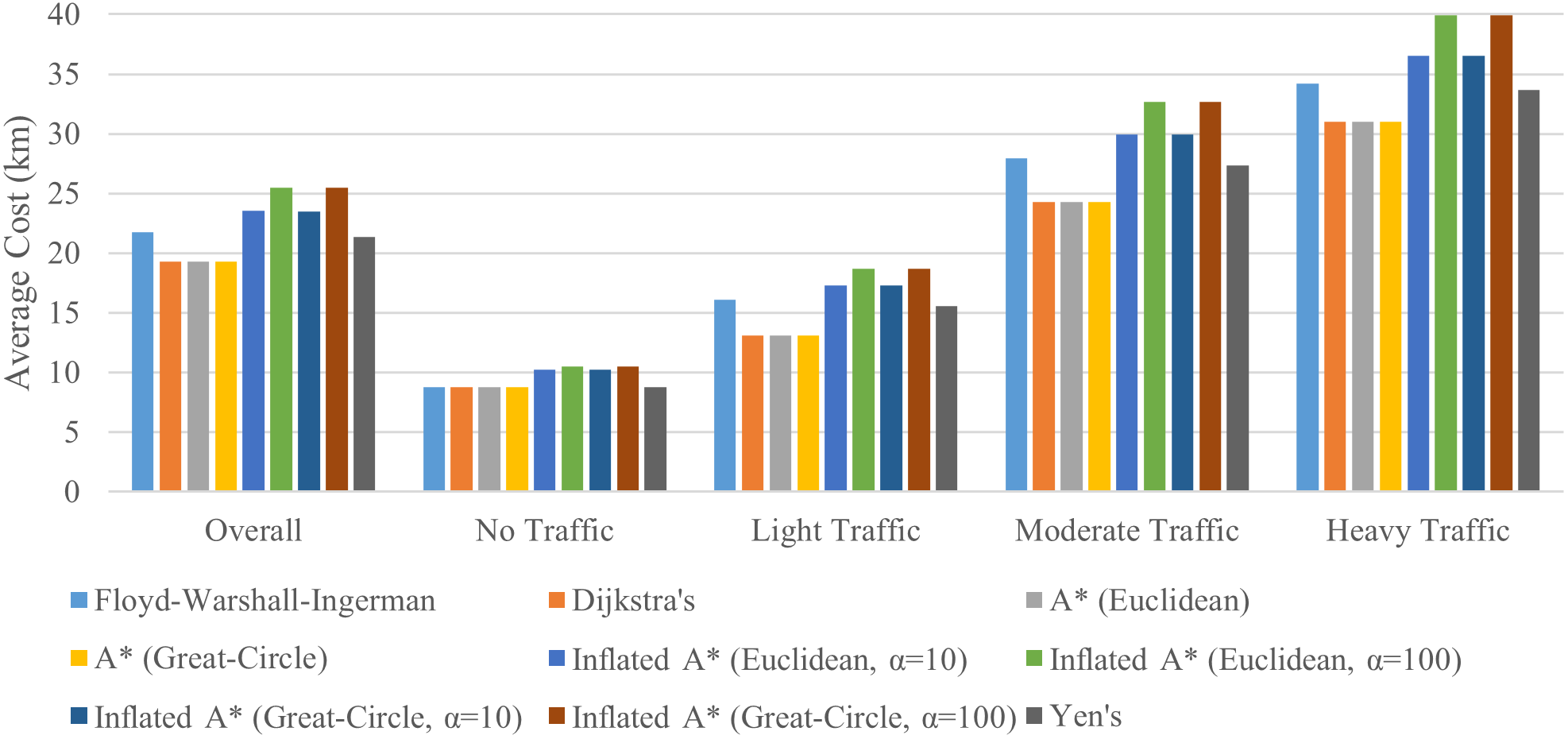}
    \caption{Average cost breakdown by trial traffic conditions with traffic-weighted distance as edge cost, ranging from no traffic to heavy traffic.}
    \label{fig:bytraffic}
\end{figure*}

Fig. \ref{fig:bytraffic} illustrates a further breakdown of average solution cost (i.e., traffic-weighted distance) by traffic condition trial category for all algorithms. It can be observed that average solution costs are lowest for the trials with no traffic and highest for the trials with heavy traffic. It is also visible that Dijkstra's algorithm and both A* searches (i.e., with Euclidean or great-circle distance) have the lowest average costs, which are identical, and that the inflated A* searches have the highest average costs.

\begin{figure*}
    \centering
    \includegraphics[width=\textwidth]{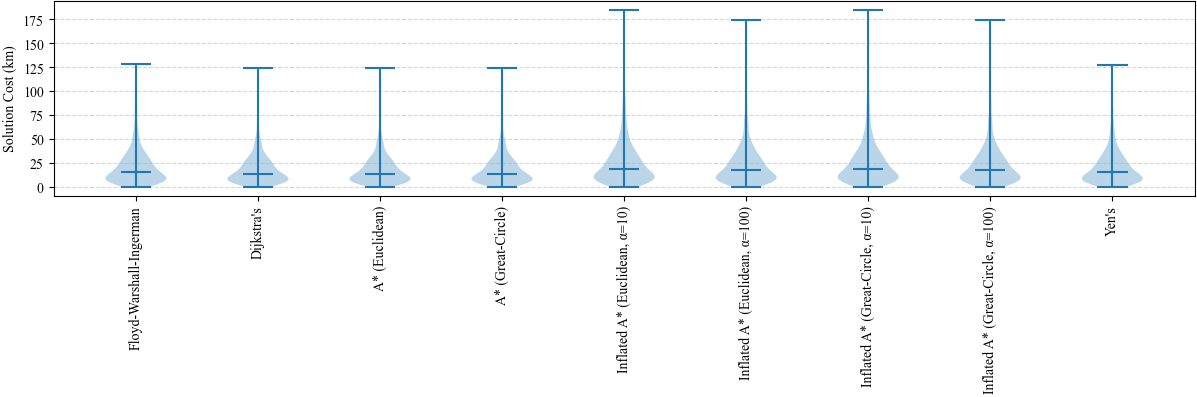}
    \caption{Violin plot showing the median, minimum, maximum, and distribution of solution costs with traffic-weighted distance as edge cost.}
    \label{fig:dist}
\end{figure*}

In addition to the average statistic being reported for solution cost (i.e., traffic-weighted distance), Fig. \ref{fig:dist} illustrates the distributions and more statistics, including the median, minimum, and maximum, through violin plots. The distributions are skewed, indicating that most trials had smaller solution costs and the larger costs are outliers. It is clear that the inflated A* searches resulted in much higher solution costs. The differences between the rest of the algorithms are only slightly noticeable. Further, due to the skewed distributions, the median costs are smaller than the average costs.

\begin{figure*}
    \centering
    \includegraphics[width=\textwidth]{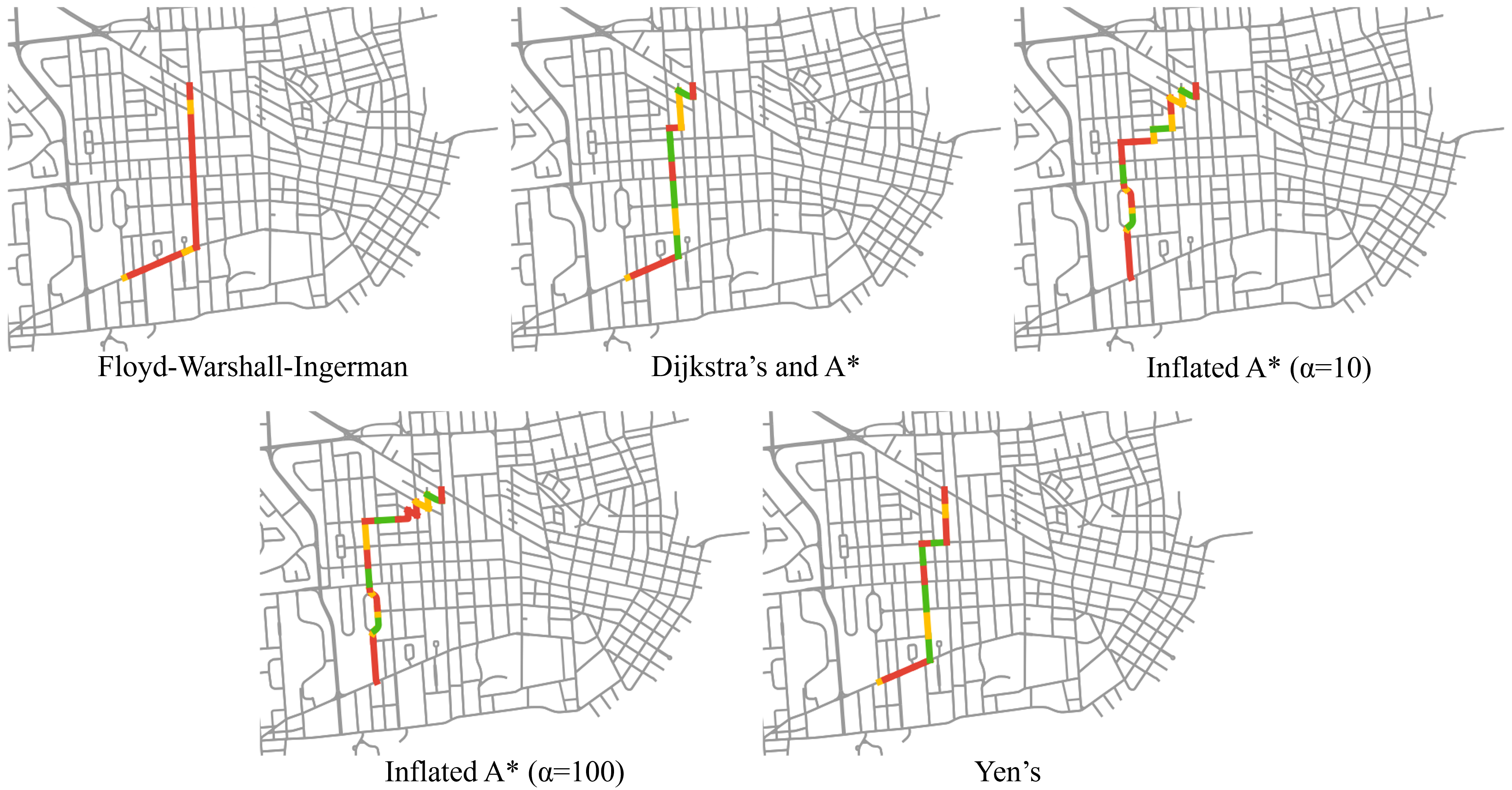}
    \caption{Traffic conditions and paths in downtown Kingston for an experiment trial with heavy traffic, with green representing roads with low traffic, orange representing roads with medium traffic, and red representing roads with high traffic. The start location is the intersection of Union St and College St (the Southern point) and the target destination is the intersection of Princess St and Victoria St (the Northern point).}
    \label{fig:paths}
\end{figure*}

The paths and traffic conditions for a heavy traffic trial are shown in Fig. \ref{fig:paths}. The start location is the intersection of Union St and College St and the target destination is the intersection of Princess St and Victoria St. The shortest distance path was found by the Floyd-Warshall-Ingerman algorithm. Dijkstra's algorithm, A* search, and Yen's algorithm slightly diverged from this path to account for traffic. Inflated A* significantly strayed away from the other paths, resulting in higher costs.

\subsection{Statistical Analysis}

To verify statistical significance of the comparison, multiple statistical tests were conducted for the nine algorithms reported in Table \ref{tab:results1}. The test was run on the solution costs of the 995 trials that had a solution. The analysis of variance (ANOVA) p-value between all sets of costs is less than 0.05. Further, the vast majority (77/81) of paired two-tailed t-test p-values between the costs of each pair of algorithms are less than 0.05, except the uninflated Single-Query Search algorithms with each other (i.e., Dijkstra's, Euclidean, and great-circle) and the inflated versions with each other (i.e., Euclidean and great-circle with $\alpha$=10 and Euclidean and great-circle with $\alpha$=100), as expected. Finally, the same observation was found with the Wilcoxon signed-rank test, which is a non-parametric version of the paired t-test.

\subsection{Incorporating Speed Limit into Edge Cost}

Although most speed limits were missing and our imputations relied on assumptions that may not necessarily be accurate, the same experiments were run one more time with speed limits incorporated into the edge costs to demonstrate the importance of including this factor. Table \ref{tab:results3} presents the evaluation metrics for all algorithms, where edge costs were calculated by multiplying the traffic weight by the distance divided by the speed limit. Most metrics follow the same trends as before, with one key exception. For A* search, the heuristic (i.e., Euclidean or great-circle distance) was modified to match the units being used to represent cost. Previously, the units were distance since the cost was traffic-weighted distance. However, the units here are time since the cost is traffic-weighted distance divided by the speed limit. Therefore, the Euclidean or great-circle distance was divided by the average speed limit to get a heuristic estimate of closeness from a vertex to the goal. The main issue with this approach is that this is not an admissible heuristic since it is possible to overestimate the true cost, resulting in slightly sub-optimal average costs compared to Dijkstra's algorithm. In the future, this can be avoid by dividing by the maximum possible speed limit instead of the average.

\begin{table*}
\caption{Performance metrics for all algorithms across the three approaches with traffic and speed limit-weighted distance as edge cost. **Only partial preprocessing was conducted for Yen's algorithm.}
\label{tab:results3}
\centering
\begin{tabular}{|c|c|c|c|c|c|}
\hline
Approach & Algorithm & \makecell[c]{Preprocessing \\ Time (min)} & \makecell[c]{Average \\ Runtime (s)} & \makecell[c]{Average \\ Cost (min)} & \makecell[c]{Average \# of \\ Vertices Expanded} \\ \hline
Multi-Query Lookup & Floyd-Warshall-Ingerman & 274.11 & \textbf{0.000029} & 20.46 & - \\ \hline
\multirow{7}{*}{Single-Query Search}
 & Dijkstra's & \textbf{0.00} & 2.44 & \textbf{18.65} & 1633 \\ \cline{2-6}
 & A* (Euclidean) & 15.63 & 2.31 & 18.70 & 890 \\ \cline{2-6}
 & A* (Great-Circle) & 18.57 & 2.33 & 18.70 & 892 \\ \cline{2-6}
 & Inflated A* (Euclidean, $\alpha$=10) & 15.63 & 0.22 & 23.69 & 64 \\ \cline{2-6}
 & Inflated A* (Euclidean, $\alpha$=100) & 15.63 & 0.20 & 25.75 & \textbf{63} \\ \cline{2-6}
 & Inflated A* (Great-Circle, $\alpha$=10) & 18.57 & 0.22 & 23.68 & 64 \\ \cline{2-6}
 & Inflated A* (Great-Circle, $\alpha$=100) & 18.57 & 0.21 & 25.75 & \textbf{63} \\ \hline
K Shortest Paths & Yen's & 3750.59** & 0.042 & 20.10 & - \\ \hline
\end{tabular}
\end{table*}

Table \ref{tab:results4} shows a direct comparison of the average estimated arrival times when using each approach to calculating edge costs. For most algorithms, incorporating both traffic and speed limits (V2) results in faster arrivals than only considering traffic (V1).

\begin{table*}
\caption{Comparison of average estimated arrival time using traffic-weighted distance only (V1) versus traffic and speed limit-weighted distance (V2).}
\label{tab:results4}
\centering
\begin{tabular}{|c|c|c|c|}
\hline
Approach & Algorithm & \makecell[c]{Average Estimated \\ Arrival Time - V1 (min)} & \makecell[c]{Average Estimated \\ Arrival Time - V2 (min)} \\ \hline
Multi-Query Lookup & Floyd-Warshall-Ingerman & 21.89 & 20.46 \\ \hline
\multirow{7}{*}{Single-Query Search}
 & Dijkstra's & \textbf{19.58} & \textbf{18.65} \\ \cline{2-4}
 & A* (Euclidean) & \textbf{19.58} & \textbf{18.70} \\ \cline{2-4}
 & A* (Great-Circle) & \textbf{19.58} & \textbf{18.70} \\ \cline{2-4}
 & Inflated A* (Euclidean, $\alpha$=10) & 23.83 & 23.69 \\ \cline{2-4}
 & Inflated A* (Euclidean, $\alpha$=100) & 25.74 & 25.75 \\ \cline{2-4}
 & Inflated A* (Great-Circle, $\alpha$=10) & 23.80 & 23.68 \\ \cline{2-4}
 & Inflated A* (Great-Circle, $\alpha$=100) & 25.74 & 25.75 \\ \hline
K Shortest Paths & Yen's & 21.44 & 20.10 \\ \hline
\end{tabular}
\end{table*}


\section{DISCUSSION}

Overall, this study demonstrated the tradeoffs between three algorithmic approaches to traffic-aware road navigation. Single-query search with Dijkstra's algorithm required the least amount of preprocessing and resulted in the most optimal paths, while multi-query lookup with the Floyd-Warshall-Ingerman required the least amount of real-time computation. K shortest paths with Yen's algorithm provided a balance between the other two approaches in terms of runtime and solution cost, but needed the most preprocessing computation. Inflated A* performed the worst in terms of solution cost, underperforming even the traffic-unaware approach.

The choice between which approach and algorithm to use depends on factors such as how frequently changes can be introduced to a road network (changes such as new roads and road closures due to construction, accidents, or weather conditions may require the time-consuming preprocessing step to be repeated), whether paths will be found locally on user devices or on a server (runtime would be a more significant consideration if a server receives many simultaneous requests), and how optimal users expect the directions to be. In small cities with minimal traffic or that have mild rush hours, the Floyd-Warshall-Ingerman approach is the least computationally expensive. In highly dynamic road networks, running a Dijkstra's algorithm or A* search for each user request would be the most feasible approach. In cities with a large user base and a fairly developed road network, investing the time and computational resources into running the Yen's algorithm preprocessing stage with a large K (to account for possible network changes) can prove beneficial in the long term. Increasing the value of K would improve optimality but increase the preprocessing time and runtime.

For A* search, the difference between using the Euclidean distance versus the great-circle distance as the heuristic is minimal. However, using the Euclidean distance resulted in less preprocessing, a faster average runtime, and a slightly smaller average number of nodes expanded. The great-circle distance measures distance between two points along the Earth's curvature and is larger than the straight-line Euclidean distance which assumes someone can move below the Earth's surface. Although the great-circle distance was expected to be a more precise heuristic and thus result in faster searches, the difference between both distance metrics is minimal for small distances (e.g., within Kingston).

The algorithms implemented here find paths between intersections. For deployment, a user can start and end at any coordinate. Therefore, an additional step would be needed to find the nearest intersections to the start and goal and instruct users to navigate from or to them.

Although traffic can change after the path is found, this was not accounted for in the scope of this project. However, it would be possible to sense if traffic significantly changes, or a road closure occurs, and re-plan from the nearest intersection to the goal.

\subsection{Hypotheses Revisited}

Most of our previous hypotheses from the the project proposal and interim presentation were correct. As expected, all algorithms found  valid routes, and the differences between them lie in the trade-offs between speed and optimality. Below is a summary of our hypotheses and results per metric:

\begin{itemize}
    \item \textbf{Preprocessing Overhead:} We hypothesized that the Floyd-Warshall-Ingerman algorithm and Yen's algorithm would take a long time find the optimal path or top K paths between all start-goal intersection pairs, and that A* search will take no time because the initialization step is not applicable. This hypothesis was mostly supported, but we underestimated that Yen's algorithm would take a much longer time than the Floyd-Warshall-Ingerman algorithm. We also decided to run a small preprocessing step to generate a heuristic lookup table for A* and save time during runtime execution.
    \item \textbf{Real-Time Computation Effort:} We hypothesized that the Floyd-Warshall-Ingerman approach would take minimal time in real time to perform a simple traffic-unaware path lookup, that Yen's algorithm approach would take a small amount of time to iterate over the top K paths and find the fastest one given traffic conditions, and that A* search would take the longest time because it must compute the optimal path from scratch. This hypothesis was fully supported by the experimental results.
    \item \textbf{Optimality:} We hypothesized that the Floyd-Warshall-Ingerman approach would result in the highest solution cost since it does not take into account traffic conditions, that A* would result in optimal solutions, and that the Yen's algorithm approach can result in sub-optimal solutions. This hypothesis was mostly upheld, but the Yen's algorithm approach did on average result in sub-optimal solutions that were notably less optimal than with A* search.
\end{itemize}

\subsection{Future Work}

Future work can focus on exploring other ways of integrating traffic information (e.g., continuous measurement as opposed to fixed traffic weights) and improving code efficiency and parallelizing the preprocessing stage to speed up the implementation. The speed limit imputation can also be improved since the assumptions that were made here are not fully accurate. For example, Kingston is reducing most neighbourhood speed limits from 50 km/h to 40 km/h \footnote{https://getinvolved.cityofkingston.ca/neighbourhood-area-speed-limits}, so not all missing speed limits for residential roads should be imputed as 50 km/h. This may require a field visit, crowdsourcing data, or searching for other sources, such as the Google Maps API which is a paid service. Finally, other heuristics can be explored since with the Euclidean or great-circle distance, the destination may be very close physically but difficult to reach due to the road network structure. An alternative could be to use the actual travel distance or time as the heuristic, but this would require a preprocessing step to find the travel distance or time between each pair of vertices.




\section{CONCLUSION}

This study presented an extensive comparison between three approaches to road navigation under a variety of traffic conditions in Kingston's road network. Two methods of weighting distance were tested for edge costs: traffic-weighted distance and distance weighted by both traffic and speed limit. Each approach's advantages and practical considerations for deployment were thoroughly discussed.


\addtolength{\textheight}{-12cm}  

\bibliographystyle{IEEEtran}
\bibliography{references}

@article{boeing2025modeling,
  title={{Modeling and Analyzing Urban Networks and Amenities with OSMnx}},
  author={Boeing, Geoff},
  journal={Geographical Analysis},
  year={2025},
  publisher={Wiley Online Library}
}

@article{warshall1962theorem,
  title={{A Theorem on Boolean Matrices}},
  author={Warshall, Stephen},
  journal={Journal of the ACM (JACM)},
  volume={9},
  number={1},
  pages={11--12},
  year={1962},
  publisher={ACM New York, NY, USA}
}

@article{floyd1962algorithm,
  title={{Algorithm 97: Shortest Path}},
  author={Floyd, Robert W},
  journal={Communications of the ACM},
  volume={5},
  number={6},
  pages={345--345},
  year={1962},
  publisher={ACM New York, NY, USA}
}

@article{ingerman1962algorithm,
  title={{Algorithm 141: Path Matrix}},
  author={Ingerman, Peter Zilahy},
  journal={Communications of the ACM},
  volume={5},
  number={11},
  pages={556},
  year={1962},
  publisher={ACM New York, NY, USA}
}

@article{dijkstra1959note,
  title = {A note on two problems in connexion with graphs},
  author = {Dijkstra, E. W.},
  journal = {Numerische Mathematik},
  year = {1959},
  publisher = {Springer-Verlag},
  volume = {1},
  number = {1},
  pages = {269–271},
}

@article{hart1968formal,
  title={{A Formal Basis for the Heuristic Determination of Minimum Cost Paths}},
  author={Hart, Peter E and Nilsson, Nils J and Raphael, Bertram},
  journal={IEEE Transactions on Systems Science and Cybernetics},
  volume={4},
  number={2},
  pages={100--107},
  year={1968},
  publisher={IEEE}
}

@article{yen1971finding,
  title={{Finding the K Shortest Loopless Paths in a Network}},
  author={Yen, Jin Y},
  journal={management Science},
  volume={17},
  number={11},
  pages={712--716},
  year={1971},
  publisher={Informs}
}

@article{cohn2009real,
  title={Real-time traffic information and navigation: An operational system},
  author={Cohn, Nick},
  journal={Transportation Research Record},
  volume={2129},
  number={1},
  pages={129--135},
  year={2009},
  publisher={SAGE Publications Sage CA: Los Angeles, CA}
}

@inproceedings{shang2013finding,
  title={Finding traffic-aware fastest paths in spatial networks},
  author={Shang, Shuo and Lu, Hua and Pedersen, Torben Bach and Xie, Xike},
  booktitle={International Symposium on Spatial and Temporal Databases},
  pages={128--145},
  year={2013},
  organization={Springer}
}

@inproceedings{xu2012traffic,
  title={Traffic aware route planning in dynamic road networks},
  author={Xu, Jiajie and Guo, Limin and Ding, Zhiming and Sun, Xiling and Liu, Chengfei},
  booktitle={International Conference on Database Systems for Advanced Applications},
  pages={576--591},
  year={2012},
  organization={Springer}
}

\end{document}